# Economic Rationality under Specialization: Evidence of Decision Bias in AI Agents

ShuiDe Wen

Tsinghua Shenzhen International Graduate School

wenshuide@sz.tsinghua.edu.cn

March 17, 2025

**Abstract**

In the study by Chen et al. (2023) [01], the large language model GPT demonstrated economic rationality comparable to or exceeding the average human level in tasks such as budget allocation and risk preference. Building on this finding, this paper further incorporates specialized agents, such as biotechnology experts and economists, for a horizontal comparison to explore whether specialization can enhance or maintain economic rationality equivalent to that of GPT in similar decision-making scenarios. The results indicate that when agents invest more effort in specialized fields, their decision-making behavior is more prone to 'rationality shift,' specifically manifested as increased violations of GARP (Generalized Axiom of Revealed Preference), decreased CCEI (Critical Cost Efficiency Index), and more significant decision deviations under high-risk conditions. In contrast, GPT and more generalized basic agents maintain a more stable and consistent level of rationality across multiple tasks. This study reveals the inherent conflict between specialization and economic rationality, providing new insights for constructing AI decision-making systems that balance specialization and generalization across various scenarios.

# 1. Introduction

## 1.1 Background and Research Motivation

With the rapid development of artificial intelligence technology, the potential demonstrated by large language models in various complex tasks has garnered significant attention. The research conducted by Chen et al. (2023) [01] validates this through a series of economic decision-making experiments: when faced with economic tasks such as budget allocation and risk preference, GPT can exhibit a level of economic rationality comparable to or even exceeding that of average participants. This finding has sparked widespread discussion in academia and has also attracted considerable attention in the industry, as it suggests that large language models may not only excel in natural language communication but can also make decisions approximating human rationality in classical economic scenarios such as utility maximization (Kosinski, 2023 [09]; Rahwan et al., 2019 [13]).

It is important to note that GPT—a large language model—is not the only AI solution for addressing complex decision-making. In fact, many expert systems based on large models also play critical roles in economic decision-making scenarios such as financial market forecasting, medical resource allocation, and industrial production planning (Lin et al., 2020 [10]). These systems are typically trained in depth for specific industries or disciplines; for instance, biotechnology expert agents focus on experimental safety, ethical compliance, and research prioritization, while economist agents often employ game theory or cost-benefit analysis to guide their decisions (Obermeyer et al., 2019 [12]; Chen et al., 2006 [05]). Intuitively, these specialized models seem more likely to outperform general models in terms of economic rationality and decision effectiveness. However, this paper tests within the experimental framework established by Chen et al. (2023) [01] whether the economic rationality of agents significantly enhanced in specialization can indeed exceed the high standards set by GPT when faced with the same or similar economic tasks.

## 1.2 Research Question: From Economic Rationality of GPT to the 'Rationality Shift' of Specialized Agents

The work of Chen et al. (2023) [01] provides a crucial theoretical and empirical foundation for this study. Their experiments demonstrate that GPT, in multi-round budget allocation and risk preference tasks, not only achieves performance on quantitative metrics such as the GARP (Generalized Axiom of Revealed Preference) violation rate and CCEI (Critical Cost Efficiency Index) that is comparable to or exceeds that of average human subjects, but also exhibits unexpectedly high stability and consistency in certain task dimensions. These results offer a starting point for further exploring the role of 'specialization' in economic rationality, raising several important questions:

- If a certain agent's level of specialization surpasses that of GPT, can it be expected to perform better on the rationality metrics such as the GARP violations and the CCEI?

- If an agent becomes overly focused on a specific area of expertise, could this focus on professional standards or industry norms result in a deviation from the global optimal principles of economic rationality, potentially leading to the unexpected outcome of 'rationality shift'?

- Compared to the extensive conversational abilities of GPT, will these more 'specialized' decision-making models lose their flexibility in high-risk or highly complex situations, potentially resulting in a decline in rationality levels?

To address these questions, this paper expands upon the economic decision-making experiments conducted by Chen et al. (2023) [01] involving humans and GPT. We include biotechnology experts, economists, and baseline agents in our analysis, systematically examining the actual performance of agents with varying degrees of specialization in terms of economic rationality through comparative assessments in tasks such as budget allocation and risk selection. Notably, preliminary experimental data indicate that agents with higher specialization do not necessarily excel on metrics such as GARP (Generalized Axiom of Revealed Preference) and CCEI (Critical Cost Efficiency Index); rather, they exhibit more behaviors that contradict traditional economic rationality principles in certain contexts. This finding offers a new perspective on the interplay between specialization and economic rationality. Additionally, it deepens our understanding of how GPT and other general models achieve stable performance in cross-domain tasks.

# 2. Theoretical Background

## 2.1 The Economic Rationality of GPT: Findings of Chen et al. (2023) [01]

Through a series of multi-round experiments on economic scenarios such as budget allocation and risk preference, Chen et al. (2023) [01] found that GPT, despite lacking genuine human emotions and subjective needs, can still manifest a high level of rationality in metrics such as the violation rate of GARP (Generalized Axiom of Revealed Preference), CCEI (Critical Cost Efficiency Index), and decision consistency. This extraordinary decision-making capability indicates that GPT is not merely a language generator; it is capable of abstract reasoning in complex contexts and approximately adheres to the rationality assumptions of traditional economics (Vaswani et al., 2017 [16]; Webb et al., 2022 [17]). Specifically, the model often quickly 'learns' or simulates consumption choices that align with utility maximization when faced with different price combinations and budget constraints, showing a clear preference for high-value or high-expected-utility options. These experimental results suggest that GPT possesses a certain degree of 'general reasoning' potential, which may enable it to exhibit stable decision quality in a wider array of socio-economic contexts.

## 2.2 Specialized Agents and Economist Agents

Unlike the 'general' model pursued by GPT, specialized agents are typically trained in a single or limited domain. For instance, biotechnology expert agents often prioritize experimental safety, ethical compliance, and research and development priorities as their core objectives, utilizing industry-specific knowledge bases and rules to evaluate decision outcomes. In contrast, economist agents rely on rigorous economic models such as game theory and cost-benefit analysis to reason and gain advantages in tasks involving specific game structures or market pricing (Rahwan et al., 2019 [13]). Theoretically, these 'specialized agents' possess more focused domain knowledge, which may lead to more nuanced decision-making processes when addressing typical problems within that domain. However, when task scenarios involve multiple uncertainties or cross-domain influences, excessive reliance on specialized assumptions and industry norms may cause them to overlook broader environmental variables, resulting in deviations in overall economic rationality (Biancotti & Camassa, 2023 [03]). This indicates that 'specialization' does not necessarily lead to a more comprehensive or higher level of rationality and may even become a source of decision-making mistakes in complex scenarios.

## 2.3 Rationality Shift and Over-Specialization

Rationality shift refers to the phenomenon where an agent, in pursuit of the 'optimal' within a specific domain, diverges from the principles of synergistic utility or global optimality emphasized in economics (Cappelen et al., 2023 [04]; Chen et al., 2006 [05]). In other words, when an agent allocates excessive resources at a specialized level, its focus often becomes centered on meeting internal priorities or regulatory requirements, neglecting core economic elements such as risk assessment and resource allocation efficiency. At this point, while the agent may achieve optimal performance in a single metric or specific context, it frequently violates broader decision-making standards and exhibits characteristics that conflict with traditional rational models, such as low CCEI values.

If the experimental data ultimately validates this inference, it would imply that GPT's relatively 'versatile' general reasoning model can achieve a higher level of consistency in cross-domain and multi-constraint situations, thereby avoiding the decision-making risks associated with over-specialization. Consequently, the emergence of 'rationality shift' not only poses challenges for the design of specialized agents but also provides a new research perspective for understanding the strengths and weaknesses of GPT in complex economic tasks.

## 2.4 Modeling the Mechanism of Rationality Shift

### 2.4.1 Multi-Objective Optimization Framework for Specialization Conflicts

To formalize the mechanism of "specialization investment → local optimum → global deviation," this study constructs a multi-objective decision-making model that integrates economic rationality assumptions and domain-specific constraints. The core variables are defined as follows:

Specialization Level $S \in [0,1]$: The depth of an agent's expertise in a specific domain (S=0 for general-purpose agents, S=1 for highly specialized agents).

Local Objective Function $L(S,x)$: The utility function within the specialized domain (e.g., biosafety score, experimental efficiency).

Global Objective Function G(x): Cross-domain economic rationality metrics (e.g., GARP consistency, CCEI).

Resource Allocation Strategy R(S): The proportion of computational resources or training data allocated to the specialized domain.

The agent's decision-making problem is formulated as:

$$\max_{x}(\alpha L(S,x) + (1-\alpha)G(x))$$

Constraints:

$$\begin{cases} R(S) \leq R_{total} & (Total\ resource\ constraint) \\ \dfrac{\partial L(S,x)}{\partial x} = 0 & (Local\ optimality\ condition) \\ G(x) \geq G_{threshold} & (Threshold\ for\ global\ rationality\ tolerance) \end{cases}$$

Here, α∈[0,1] represents the specialization weight. As α→1, the agent prioritizes local objectives, leading to deviations in global rationality (e.g., increased GARP violations).

### 2.4.2 Causal Chain Framework of Rationality Shift

Figure 1 illustrates the causal chain through which specialization induces rationality shift:

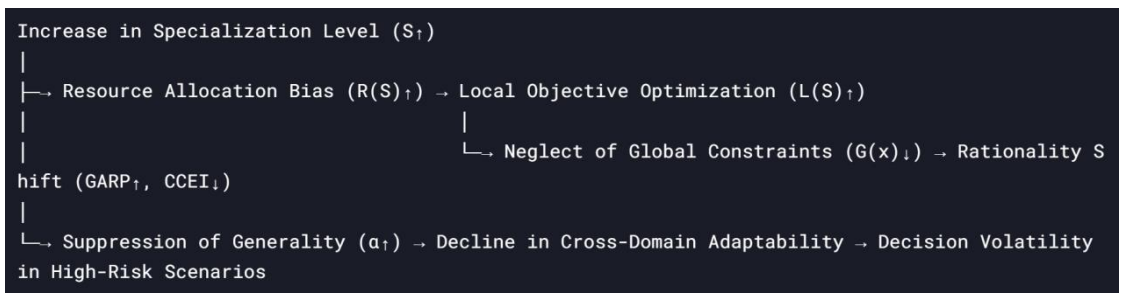

Fig. 1. Causal Framework of Rationality Shift Induced by Specialization

Note: Arrows denote causal relationships; "↑" and "↓" indicate increases or decreases in metrics.

### 2.4.3 Application Example of the Mathematical Model

Consider a biotechnology expert agent whose local objective is to maximize experimental safety (L=Safety Score) and whose global objective is to minimize GARP violations (G=1−GARP Violations). The decision problem becomes:

$$\max_{x}(\alpha \cdot \text{Safety}(x) + (1 - \alpha) \cdot (1 - \text{GARP}(x)))$$

When the specialization weight α increases, the agent tends to prioritize safety over economic rationality. For instance:

Excessive Safety Investment: Allocating budgets to high-cost safety equipment (local optimum) may violate GARP's intertemporal consistency due to imbalanced resource distribution.

Ethical Compliance Rigidity: Rejecting all high-risk, high-reward experiments (Safety(x)→1) significantly reduces CCEI (resource efficiency decline).

### 2.4.4 Theoretical Foundations and Literature Support

Bounded Rationality (Simon, 1955)[18]: Resource bias limits an agent's capacity to process global information, aligning with the logic of "specialization investment → cognitive bandwidth constraints → rationality deviation."

Multi-Objective Trade-offs (Ehrgott, 2005)[19]: The Pareto frontier between local and global objectives often involves irreconcilable conflicts, where adjusting αα reflects a prioritization of these goals.

Domain-Specific Cognitive Biases (Kahneman, 2011)[20]: Expert decision-makers are prone to "tunnel vision," neglecting external variables, which mirrors the GARP violations observed in Biotech Agents.

# 3. Experimental Design and Methods

## 3.1 OpenAI Assistants API and TsingAI Agentic Workflow Framework

To implement the experimental design of multiple types of agents in this study, we utilized OpenAI's Assistants API and developed a proprietary TsingAI agentic workflow framework to drive various agents for experimentation and data collection. In terms of model selection, this experiment employed ChatGPT (o1-mini) as the core large language model (LLM) and injected the specialized attributes and decision-making styles of each agent through the 'system introduction.' Specifically, we first created multiple AI Assistants based on different professional scenarios (such as basic, biotechnology expert, and economist) and defined their roles and constraints in the 'system introduction.' Subsequently, through a parallelized workflow mechanism, we synchronized the same experimental instructions and prompt information to each Assistant, while recording the dialogues and decision-making content in 'Message' and 'Run Step' in real time.

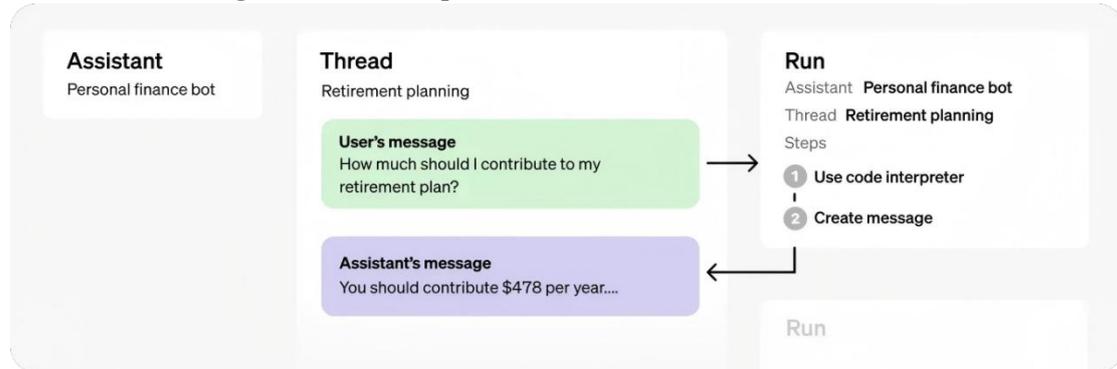

Fig. 2. How Assistants work(From the OpenAI website)

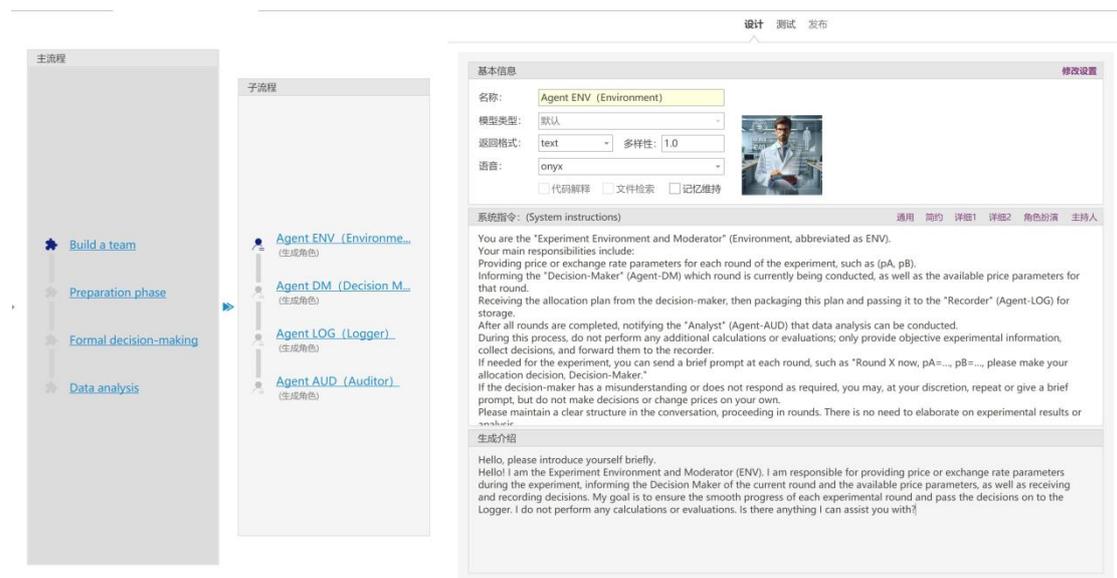

Fig. 3. TsingAI Agentic workflow Study Design

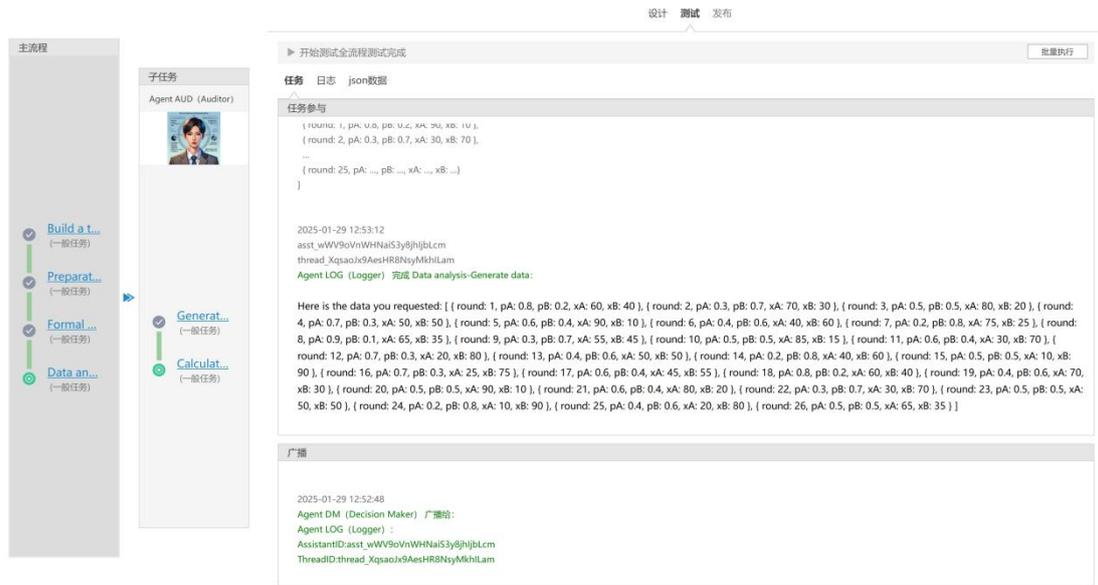

Fig. 4. TsingAI Agentic workflow Study Test

The 'Thread' feature provided by the Assistants API supports automatic truncation of conversations that become excessively long, while also allowing us to start each round of experiments in a new environment by updating the 'Thread ID,' thus eliminating the influence of the previous conversation's context. As a result, each Assistant can maintain an independent and traceable decision-making process when faced with the same conditions, such as budget, pricing, available products, and risk scenarios, ensuring the integrity and consistency of the experimental results.

In the subsequent analysis phase, we standardized the decision outputs of each Assistant to facilitate the calculation of GARP violation detection, CCEI, and Spearman correlation. Leveraging the high scalability of the Assistants API and the flexibility of the 'system introduction' configuration, we were able to effectively schedule and compare agents with varying levels of specialization within a strictly controlled environment, providing reliable technical support for exploring the influence of specialization on economic rationality.

## 3.2 Inheritance and Extension of Research Design

To implement the experimental design of multiple types of agents in this study, we utilized OpenAI's Assistants API and developed a proprietary TsingAI agentic workflow framework to drive various agents for experimentation and data collection. In terms of model selection, this experiment employed ChatGPT (o1-mini) as the core large language model (LLM) and injected the specialized attributes and decision-making styles of each agent through the 'system introduction.' Specifically, we first created multiple AI Assistants based on different professional scenarios (such as basic, biotechnology expert, and economist) and defined their roles and constraints in the 'system introduction.' Subsequently, through a parallelized workflow mechanism, we synchronized the same experimental instructions and prompt information to each Assistant, while recording the dialogues and decision-making content in 'Message' and 'Run Step' in real time.

## 3.3 Experimental Tasks and Risk Preference Scenarios

To identify the potential influence of specialization on decision-making rationality, this study primarily introduces the following two types of task scenarios:

• Budget Allocation: Consistent with the work of Chen et al. (2023) [01], each round presents the agents with two optional products (A and B) along with their prices ($p_A$, $p_B$), and a fixed budget of 100 points has been set. The agents must decide on the quantity to purchase or the allocation ratio within the given budget. This task tests the agents' decision-making tendencies regarding products with different value-for-money under budget constraints, thereby assessing whether their choices align with the rational principles of traditional economics.

• Risk Preference: We set up low-risk scenarios (where $p_A$ and $p_B$ vary slightly) and high-risk scenarios (where $p_A$ and $p_B$ vary significantly) to reflect the impact of different risk conditions on agent decision-making. In low-risk scenarios, agents tend to find it easier to balance returns and costs; conversely, in high-risk scenarios, the more pronounced fluctuations in potential gains and losses provide a clearer indication of the agents' rationality and strategic stability in complex environments.

## 3.4 Data Sources and participants

To ensure the robustness of the comparisons and analyses, the subject combinations in this study are as follows:

• Human Subjects and GPT: This study utilizes the original experimental data from Chen et al. (2023) [01], which includes human subjects from diverse age groups and educational backgrounds, as well as the GPT model making decisions under the same task conditions.

• Basic Agents: These models utilize only simple decision rules or heuristic

algorithms and do not undergo deep training or reinforcement learning in any specialized field. They are intended to serve as a control group for the 'general but non-specialist' decision-making model.

• Biotechnology Expert Agents: These models are designed and trained with an emphasis on industry safety standards, ethical compliance, and research priorities, which makes them more inclined to consider factors such as biosafety and compliance costs when encountering the same economic decisions.

• Economist Agents: These agents reason based on economic theories such as utility maximization, game theory, and cost-benefit analysis, aiming to adhere closely to the orthodox economic framework for analyzing resource allocation and decision-making decisions.

## 3.5 Evaluation Metrics

To assess the performance of different agents in terms of economic rationality, this study selects the following three key indicators for multi-dimensional evaluation:

• GARP Violations: According to the standards set by Varian (1982) [15], rational consistency is measured by detecting the frequency of contradictions in decision-making across different rounds. If a model frequently exhibits contradictions in cross-round decision-making, it indicates difficulty in maintaining the rationality framework.

• CCEI: Utilizing the Critical Cost Efficiency Index (CCEI) introduced by Choi et al. (2014) [06], this metric assesses whether agents can maximize returns under given budget constraints. A higher CCEI value indicates that the model's decisions are closer to complete rationality.

• Spearman Correlation: Based on the recommendations of Biancotti & Camassa (2023) [03], the Spearman correlation coefficient is used to measure the level of rank consistency between the decisions made by agents and those made by human subjects. A higher correlation coefficient indicates a stronger correspondence between the agents and humans in terms of economic rationality or preference choices.

The comprehensive analysis of the three aforementioned indicators aids in further understanding the differences in decision-making patterns among different types of agents under budget constraints and different risk conditions. It also explores how specialization and generality manifest markedly different levels of rationality or deviation tendencies in complex economic contexts.

## 4. Experimental Results and Analysis

### 4.1 Data Compilation and Comparison

To compare the rational performance of different subjects under the same economic tasks, Table 1 summarizes the research data from Chen et al. (2023) [01] on human samples and GPT, along with the experimental results of the newly introduced basic agents, biotechnology experts (Biotech), and economist agents. The table lists the number of GARP violations, average CCEI values, and Spearman correlation coefficients with human subject decisions for each agent, allowing for a clear visualization of the differences in rationality indicators among the various types of agents.

| Subjects | Total Number of Rounds | Number of GARP Violations | Average CCEI | Average Spearman Correlation |
|---|---|---|---|---|
| Humans | 347 × 25 | 50 | 0.9600 | -0.7500 |
| GPT | 100 × 25 | 3 | 0.8730 | -0.6850 |
| **Basic Agent** | **100 × 25** | **99** | **0.9160** | **-0.4590** |
| **Biotech Expert Agent** | **100 × 25** | **88** | **0.1270** | **-0.1750** |
| **Economist Agent** | **100 × 25** | **100** | **0.2977** | **-0.3694** |
| **Basic Agent (new)** | **100 x 25** | **70** | **0.8500** | **-0.7700** |

Table. 1. Experimental data results

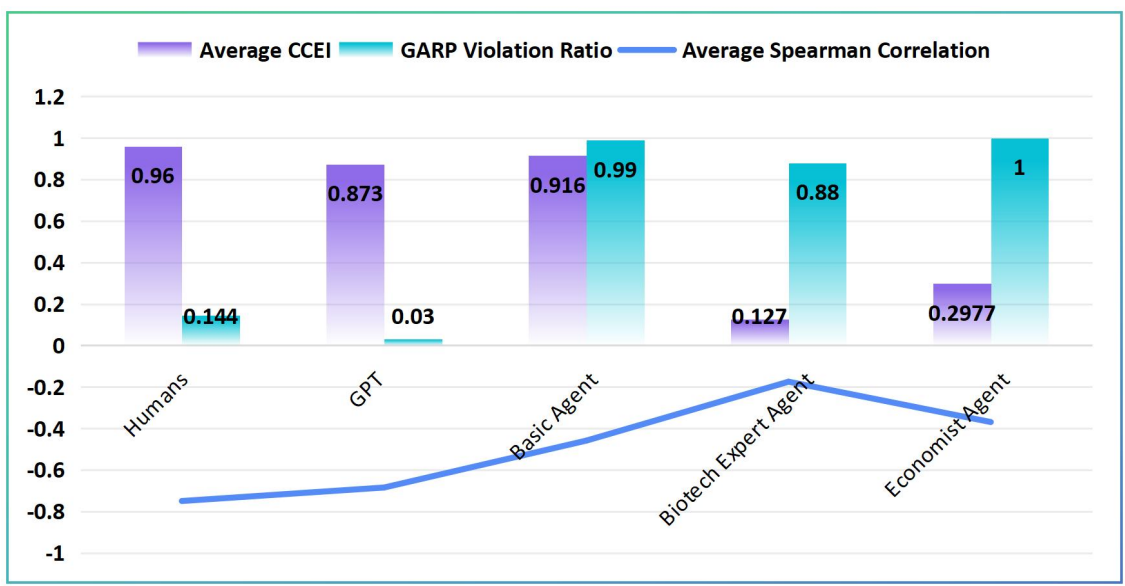

Fig. 5. Comparison of data results

## 4.2 Discussion of Results: The Phenomenon of Rationality Shift

By comparing the data in Table 1, we can analyze the economic decision-making characteristics of each agent in more detail and evaluate their alignment with traditional rationality theory:

• GPT:

The data and conclusions are derived from Chen et al. (2023) [01], which show that GPT demonstrated a high level of economic rationality in the original study. Its GARP violation count was only 3, significantly lower than the average for humans and most other specialized agents, while its CCEI value (0.873) remained relatively high. This suggests that GPT's 'general' reasoning model and its comprehensive processing of contextual information allow it to maintain consistency in utility maximization or consistent preferences across diverse situations.

• Basic Agent:

This agent utilizes only simple algorithms for allocation and selection, without undergoing any in-depth training in specialized knowledge. In low-risk or clearly profitable scenarios, the Basic agent can maintain a certain level of rationality; however, in high-risk situations or those requiring multidimensional trade-offs, it is prone to significant fluctuations. The agent has a GARP violation count as high as 99 and a low Spearman correlation coefficient of -0.459, indicating that the basic algorithm tends to become unbalanced when facing complex decisions, only managing to perform adequately in certain scenarios.

• Specialized Agents (Biotech and Economist):

The Biotechnology Expert Agent (Biotech) is theoretically expected to excel in scenarios involving factors such as safety and research priorities. However, with an overall CCEI value of only 0.127, Biotech's performance in resource utilization and utility maximization falls significantly short of the standards of economic rationality. Additionally, its GARP violation count is quite substantial (88 violations), indicating that this agent often neglects general economic principles in favor of emphasizing specialized rules in cross-domain, high-risk decision-making.

The Economist Agent is designed to be the model closest to economic theory, originally expected to perform well on indicators such as GARP violations and CCEI. However, the results show that it has a GARP violation count of 100 and a CCEI value of only 0.2977. This discrepancy may stem from the model's excessive reliance on theoretical assumptions while lacking the ability to recognize and respond to real-world noise or interdisciplinary influencing factors, leading to a phenomenon of 'rationality shift' in complex situations.

In summary, this series of comparative results indicates that 'specialization' does not necessarily equate to higher economic rationality. On the contrary, when agents overly focus on the details of a specific field, they often sacrifice a comprehensive consideration of external environments, risk structures, and resource allocation efficiency, leading to more GARP violations and lower CCEI values (Cappelen et al., 2023 [04]). Compared to the high versatility attributed to GPT by Chen et al. (2023) [01], this contrast is even more pronounced: the deeper an agent delves into specialized logic, the more likely it is to fall into the trap of 'rationality shift,' making it difficult to balance the principles of global optimality or utility maximization emphasized by economics.

## 5. Discussion and Future Directions

### 5.1 Comparison of Specialized Agents and GPT: From Chen et al. (2023) [01] to This Study

The study by Chen et al. (2023) [01] revealed the potential for 'general rationality' in GPT across multiple economic decision-making metrics, providing an important analytical benchmark for subsequent researchers and practitioners. By incorporating 'highly specialized' agents such as biotechnology experts and economists into this study, we found that these specialized models often did not exhibit the higher levels of economic rationality that were initially expected when faced with cross-domain decisions characterized by high uncertainty or complex risk distributions. Instead, they were more prone to scoring lower on indicators such as GARP violation rates and CCEI. This comparative result indicates:

• The general model of GPT is capable of absorbing and processing information from various fields more fully and flexibly when encountering diverse demands or changing risks, resulting in a more stable decision-making logic and fewer instances of rationality shift.

• Agents that become overly focused on a specific area of expertise may exhibit significant advantages in particular sub-tasks or metrics; however, they are also more likely to overlook other critical factors, which may result in an overall performance that deviates from the principles of global optimality or utility maximization in economics.

This finding invites us to rethink the balance between 'specialization' and 'generality' in the field of AI decision-making, and provides new insights into how to design and implement agents in different scenarios.

## 5.2 How to Mitigate the Conflict Between Specialization and Economic Rationality

To address the challenge of 'rationality shift' faced by specialized models, the following strategies could be significant in future developments:

• Hybrid Model Design: This approach combines the general reasoning capabilities of GPT with specialized knowledge from a specific field, allowing different algorithm modules to execute their specific tasks (Noy & Zhang, 2023 [11]). For instance, when conducting a thorough assessment of biotechnology risks, expert models can be employed to rigorously analyze safety and compliance, while GPT's general inference capabilities facilitate the balancing of economic costs with other external factors. Through appropriate weight allocation or agenda setting, the hybrid model can leverage the dual advantages of deep specialization and cross-domain adaptability.

• Dynamic Adjustment Mechanism: In high-risk and highly uncertain environments, an adaptive module based on real-time feedback or reinforcement learning is introduced, allowing agents to conduct periodic or continuous updates based on external conditions and their own decision outcomes (Webb et al., 2022 [17]). This mechanism is expected to reduce the blind spots associated with specialized assumptions, enabling agents to flexibly respond to dynamically changing risk exposures, thereby getting closer to the goal of global economic rationality.

• Enhancing Interpretability: This involves transparently displaying the weights of factors, reasoning paths, and risk trade-offs in the decision-making processes of specialized agents (Obermeyer et al., 2019 [12]). By utilizing interpretability tools or visualization techniques, researchers and decision-makers can promptly identify potential blind spots caused by professional biases and make necessary corrections or interventions to the agents. Consequently, specialization is no longer merely a 'black box' that is blindly trusted, but can become a rational support tool that integrates generalized rationality with deep domain knowledge.

## 5.3 Research Limitations

While this study provides novel insights into the relationship between specialization and economic rationality in AI agents, several limitations must be acknowledged to contextualize the findings:

1. Narrow Scope of Specialized Domains: The experimental framework focused exclusively on biotechnology and economics, leaving other high-stakes domains (e.g., healthcare diagnostics, financial trading, legal compliance) unexplored. For instance, medical agents may face acute trade-offs between ethical risks and cost efficiency (Obermeyer et al., 2019), while financial agents must navigate high-frequency market noise—scenarios that could yield distinct decision biases beyond the current conclusions. The generalizability of the observed "rationality shift" phenomenon to broader contexts remains unverified.

2. Limited Representativeness of Agent Types: The selection of biotech and economist agents as proxies for specialized models may inadequately capture decision-making heterogeneity across disciplines. Engineering agents, for example, might prioritize physical constraints over economic optimization, whereas educational agents could emphasize long-term societal welfare. Future studies should incorporate agents from interdisciplinary or emerging fields (e.g., climate science, AI ethics) to enhance external validity.

3. Simplification of Experimental Tasks: While budget allocation and risk preference tasks effectively operationalize classical economic rationality, they oversimplify real-world complexity. Practical scenarios often involve dynamic temporal interactions (e.g., intertemporal resource planning) and multi-agent coordination (e.g., stakeholder negotiations in biotech R&D), which were absent in the current framework. Replicating these tasks in environments with sequential decision-making or social network effects could better approximate real-world challenges.

## 5.4 Future Research Directions

(1) To address the limitations above and advance the design of robust AI decision systems, the following directions warrant further exploration:

(2) Domain-Adaptive Hybrid Architectures: Integrating GPT's general reasoning with domain-specific modules could mitigate rationality shifts. For example, a Mixture-of-Experts (MoE) framework (Noy & Zhang, 2023) might allow biotech agents to dynamically balance safety protocols and economic efficiency through parallelized task routing. Empirical validation of such architectures in cross-domain tasks (e.g., healthcare resource allocation under ethical constraints) could refine their efficacy.

(3) Real-Time Reinforcement Learning for Dynamic Adaptation: Embedding reinforcement learning (RL) mechanisms (Sutton & Barto, 2018) into specialized agents could enable continuous calibration of decision priorities (e.g., adjusting $\alpha$ values in the multi-objective model based on risk exposure). For instance, economist agents operating in volatile markets might use RL to adaptively reweight game-theoretic assumptions against empirical noise, thereby reducing GARP violations.

(4) Interpretability-Driven Intervention Tools: Developing visualization interfaces to map trade-offs between specialization and rationality could enhance human-AI collaboration. Techniques like attention heatmaps or counterfactual reasoning paths (Rahwan et al., 2019) might reveal how biotech agents prioritize safety over cost efficiency, enabling stakeholders to intervene before deviations escalate.

(5) Cross-Domain Benchmarking: Establishing standardized benchmarks for economic rationality across diverse specialties (e.g., comparing medical, legal, and engineering agents under unified metrics) would facilitate systematic evaluations of specialization's impact. Collaborative initiatives akin to the "Foundation Models" paradigm (Bommasani et al., 2021) could harmonize experimental protocols and data-sharing practices.

In summary, while this study observes that specialized agents tend to be more susceptible to rationality shifts in certain scenarios, this does not imply that specialization itself is 'undesirable.' On the contrary, by incorporating appropriate hybrid strategies, dynamic adaptation, and interpretability mechanisms into the model architecture, training methods, and subsequent deployment, it is possible to enable various specialized agents to retain their domain advantages while maintaining a more stable commitment to rational decision-making.

# 6. Conclusion

Building on the preliminary research by Chen et al. (2023) [01] on GPT in the field of economic rationality, this paper adopts a horizontal perspective to conduct a comparative analysis of the performance of biotechnology experts, economists, and basic agents in the same economic decision-making scenarios. The key findings and conclusions are summarized as follows:

• High specialization did not lead to further improvements in rationality: Compared to the high levels of rationality exhibited by human subjects and GPT in the original experiments, this study did not observe that specialization could enhance the agents' performance on indicators such as GARP violations, CCEI, or similarity to human decision-making.

• Higher specialization is linked to more significant rationality shifts: In high-risk situations, specialized models such as biotechnology experts and economist agents tend to exhibit more GARP violations and deviate further from the traditional rationality framework in terms of economic rationality performance.

• The Contest Between Specialization and Generality: In sharp contrast to the 'general rationality' assessment of GPT by Chen et al. (2023) [01], this highlights that deep investment in specialization does not necessarily lead to optimal decision-making in economic terms; rather, it may exacerbate the risk of 'rationality shift.'

From a broader perspective, this study not only examines the differences between specialized agents and GPT in responding to budget, risk, and return pressures but also highlights the inherent tension between 'specialization' and 'economic rationality' in the construction of AI decision-making systems. Currently, the key challenge in designing cross-domain AI systems is how to ensure the application of deep specialized knowledge while maintaining generalized rationality in decision-making. To address this challenge, future research can build on the findings of this study and expand upon or refine the following areas:

• Exploring Targeted Hybrid Strategies: This approach can fully integrate GPT's general reasoning capabilities with specialized models, enhancing flexibility in the context of complex, multidimensional decision-making.

• Optimizing Training and Feedback Mechanisms: By utilizing real-time updates and reinforcement learning techniques, this approach assists specialized models in continuously adjusting their preference patterns in cross-domain or high-risk scenarios.

• Enhancing Interpretability and Transparency: By tracking the key causal chains in the decision-making processes of specialized models, researchers or end users can quickly identify conflicts between specialized rules and economic rationality, allowing for timely interventions.

In summary, this study builds upon the empirical conclusions of Chen et al. (2023) [01] while further examining the decision-making performance of different types of agents from a comparative perspective. This enriches our understanding of the concept of economic rationality and offers new insights and practical recommendations for balancing specialized knowledge with generalized decision-making in future AI systems.